\documentclass[10pt,a4paper]{article}

\usepackage[margin=1in]{geometry}
\usepackage[T1]{fontenc}
\usepackage[utf8]{inputenc}
\usepackage{amsmath,amsfonts,amssymb}
\usepackage{booktabs}
\usepackage{microtype}
\usepackage{graphicx}
\usepackage{enumitem}
\usepackage{placeins}
\usepackage{url}
\usepackage[hidelinks]{hyperref}

\title{\large\bfseries
A Sobel-Gradient MLP Baseline for Handwritten\\ Character Recognition}

\author{Azam Nouri \\[4pt]
\small Department of Science, Technology \& Mathematics, Lincoln University}
\date{}

\begin{document}
\maketitle

\begin{abstract}
\noindent
This study examines how much handwritten-character information is retained by a deliberately simple first-order edge representation. Instead of learning spatial filters, each input image is transformed by the fixed Sobel--Feldman operator into signed horizontal and vertical derivative maps, which are independently normalized, flattened, and classified by a multilayer perceptron (MLP). The resulting model therefore separates fixed edge extraction from learned classification and provides a controlled baseline for evaluating the sufficiency of first-order image gradients. In the executed experiments, the Sobel-gradient MLP achieves \textbf{98.54\,\%} test accuracy on MNIST and \textbf{92.50\,\%} on the TensorFlow Datasets (TFDS) EMNIST Letters configuration. Macro F1 scores are 0.9853 and 0.9265, respectively. One-vs-rest ROC analysis further yields micro/macro AUC values of 0.9998/0.9998 on MNIST and 0.9987/0.9982 on EMNIST Letters. Confusion-matrix analysis shows that the remaining errors are concentrated among geometrically similar classes, especially 3/8 and 4/9 for MNIST and I/L and G/Q for EMNIST Letters. These results show that fixed first-order gradients preserve substantial class-discriminative structure, while also revealing the specific ambiguities that remain when recognition is driven by edge geometry alone.

\medskip
\noindent
\textbf{Keywords:} handwritten recognition; Sobel operator; edge detection; multilayer perceptron; MNIST; EMNIST
\end{abstract}

\section{Introduction}
\label{sec:intro}

Handwritten-character recognition (HCR) is a long-standing problem in computer vision, with applications ranging from document digitization and form processing to note transcription. Modern recognition systems are commonly built around convolutional neural networks (CNNs), which learn spatial filters directly from pixel intensities. Their success, however, can obscure a simpler representational question: how much of the information needed to distinguish handwritten characters is already exposed by a fixed first-order description of local intensity change?

For handwritten glyphs, class identity is strongly tied to stroke geometry. A character is distinguished not only by where ink appears, but also by where stroke boundaries occur and how those boundaries are oriented. First-order image derivatives provide a direct description of these local transitions. This motivates a controlled experiment in which the spatial front end is not learned: the image is transformed by the classical Sobel--Feldman operator, and a dense classifier receives only the resulting signed gradient maps.

\paragraph{Why revisit Sobel in an MLP setting?}
The Sobel--Feldman operator uses two fixed \(3\times3\) kernels to approximate horizontal and vertical derivatives while applying mild smoothing in the orthogonal direction. The transformation is deterministic, inexpensive, and directly interpretable: each input dimension supplied to the classifier corresponds to a signed local intensity change at a known image location. This makes Sobel useful not because it is a new feature extractor, but because it provides a deliberately constrained representation with which to ask how much recognition performance can be obtained from first-order edge geometry alone.

The objective of this paper is therefore not to claim that Sobel features are generally superior to learned convolutional representations. Nor does the experiment establish that Sobel preprocessing improves on raw pixels, because a matched raw-pixel MLP ablation is not included. Instead, the paper asks a narrower and more interpretable question: whether a classifier that receives only fixed first-order derivatives can still recover strong digit- and letter-level discrimination.

\paragraph{Research question.}
How much class-discriminative information is retained in signed horizontal and vertical Sobel derivatives when the downstream classifier is an all-dense MLP with no learned spatial convolution?

\paragraph{Contributions.}
\begin{enumerate}[leftmargin=1.5em]
    \item A controlled Sobel-gradient MLP baseline in which all spatial filtering is fixed and only the dense classifier is learned.
    \item A minimal and reproducible pipeline using two signed derivative channels, per-image channel normalization, a fixed MLP architecture, and no data augmentation.
    \item An empirical evaluation on MNIST and the TFDS EMNIST Letters configuration using accuracy, macro/micro precision--recall--F1, confusion matrices, and one-vs-rest ROC--AUC analysis.
    \item A class-level analysis showing where first-order edge geometry succeeds and where it remains ambiguous, particularly for geometrically similar digit and letter pairs.
    \item Reproducible code and training details to facilitate follow-up work (see Code Availability).
\end{enumerate}

\section{Related Work}
\label{sec:related}

\paragraph{Gradient descriptors for HCR.}
Before end-to-end deep feature learning became dominant, handwritten recognition systems frequently relied on manually designed contour, gradient, and directional descriptors followed by conventional classifiers. Gradient-based contour encodings and benchmark studies of handwritten digit recognition demonstrated that local edge structure can be highly informative for character discrimination~\cite{Liu2003,Srikantan1996}. The present work follows that representational tradition but asks a more specific question: whether the two raw signed Sobel derivative maps themselves can serve as the complete visual input to a dense neural classifier.

\paragraph{Edge-aware learned representations.}
Modern CNN-based systems can learn edge-like filters in their early layers, and edge-aware processing has also been incorporated explicitly into learned models~\cite{EdgeBranch2023}. In such approaches, however, spatial feature extraction remains part of the learned pipeline. Here the front end is intentionally fixed: Sobel filtering supplies the only spatial derivative representation, while all trainable parameters are confined to the MLP classifier. This separation makes the role of the input representation easier to inspect.

\paragraph{All-MLP architectures.}
All-MLP vision architectures, including MLP-Mixer~\cite{MLPMixer2021}, have shown that convolution is not the only mechanism by which image information can be processed. Those models nevertheless learn sophisticated token- and channel-mixing transformations. The present baseline is deliberately more restrictive: the model receives a flattened pair of fixed derivative maps, with no learned local spatial operator before the dense layers.

\paragraph{First-order and higher-order geometric representations.}
A related line of work studies higher-order geometric information such as planar curvature~\cite{NouriCurv2025}. The Sobel baseline considered here occupies a complementary position: rather than encoding how a contour bends, it evaluates how far first-order directional intensity changes alone can support recognition. This distinction helps separate the contribution of basic edge structure from richer geometric descriptors.

\section{Background: The Sobel Gradient}
\label{sec:sobel_background}

\subsection{First-Order Edge Representation}
\label{sec:sobel_intuitive}

An image can be viewed as a scalar intensity field \(I(x,y)\). A handwritten stroke creates rapid spatial changes in that field near its boundaries. The local derivatives \(\partial_x I\) and \(\partial_y I\) therefore provide a compact description of where intensity changes occur and in which signed direction they occur. For thin handwritten strokes, much of the visible shape is represented by these transitions around the stroke contour.

The Sobel operator estimates these derivatives with a small local stencil. One kernel responds primarily to left--right intensity change and the other to top--bottom change. Applying the pair across an image produces two derivative maps whose spatial pattern traces the orientation and polarity of character boundaries. A dense classifier can then learn combinations of those edge responses across the full image without learning the derivative filters themselves.

This representation is intentionally limited. It emphasizes local transitions and suppresses part of the absolute intensity information available in the raw image. Consequently, the experiment tests the sufficiency of a first-order edge representation rather than the full information content of the original pixels.

\subsection{Discrete \texorpdfstring{$3\times3$}{3x3} Kernels}

The Sobel--Feldman operator~\cite{Sobel1968} approximates the spatial image gradient by convolving the input \(I\) with two kernels:
\[
G_x=\!\begin{bmatrix}
-1&0&1\\[2pt]
-2&0&2\\[2pt]
-1&0&1
\end{bmatrix},
\qquad
G_y=\!\begin{bmatrix}
-1&-2&-1\\[2pt]
0&0&0\\[2pt]
1&2&1
\end{bmatrix}.
\]

Both filters perform a central difference in one direction while embedding mild smoothing with weights \(1\!:\!2\!:\!1\) along the orthogonal axis. Their integer coefficients also make the operation straightforward to implement in fixed-point or integer-oriented settings.

\subsection{Interpreting the Output}

Convolving \(I\) with \(G_x\) and \(G_y\) yields horizontal and vertical derivative maps, respectively:
\[
\partial_x I = G_x * I,\qquad
\partial_y I = G_y * I,
\]
where \(*\) denotes 2-D convolution.

From these first-order derivatives one can derive the gradient magnitude
\[
\lVert\nabla I\rVert
=
\sqrt{(\partial_x I)^2+(\partial_y I)^2}
\]
and the gradient orientation
\[
\theta
=
\operatorname{atan2}(\partial_y I,\partial_x I).
\]

In this study the derived magnitude and angle are not used as classifier inputs. Instead, the raw signed derivatives \((\partial_x I,\partial_y I)\) are retained. This choice preserves directional polarity separately along the two image axes and allows the MLP to learn task-specific combinations of positive and negative edge responses.

\section{Method}
\label{sec:method}

The experimental design intentionally keeps the representation and classifier simple: fixed Sobel derivatives \(\rightarrow\) independent channel normalization \(\rightarrow\) flattening \(\rightarrow\) MLP. There is no learned convolution, no handcrafted feature aggregation beyond the Sobel transform, and no data augmentation.

\subsection{Data Flow and Preprocessing}

Images are \(28\times28\) grayscale glyphs. Each image is passed through OpenCV's \(3\times3\) Sobel operator to obtain
\[
G_x,G_y\in\mathbb{R}^{28\times28}.
\]

Each derivative channel is min--max normalized independently, with the minimum and maximum computed for that image and channel:
\[
\widehat{G}_x
=
\frac{G_x-\min(G_x)}
{\max(G_x)-\min(G_x)+\varepsilon},
\qquad
\widehat{G}_y
=
\frac{G_y-\min(G_y)}
{\max(G_y)-\min(G_y)+\varepsilon},
\quad
\varepsilon=10^{-8}.
\]

The normalized maps are stacked as
\[
[\widehat{G}_x,\widehat{G}_y]
\in
\mathbb{R}^{2\times28\times28}
\]
and flattened to a feature vector
\[
\mathbf{x}\in\mathbb{R}^{1568}.
\]

This normalization puts the two derivative channels on a consistent numerical scale for optimization. At the same time, because the scaling is performed independently for each image and channel, absolute gradient amplitude is not preserved across examples. The model therefore primarily receives the relative spatial pattern of signed edge responses within each glyph.

\subsection{Classifier Architecture}

We use a three-hidden-layer MLP with batch normalization (BN), ReLU activations, and dropout. The output dimension is 10 for MNIST and 26 for EMNIST Letters.

\begin{table}[htbp]
\centering
\caption{MLP architecture used for the Sobel-gradient representation.}
\label{tab:architecture}
\begin{tabular}{@{}lcc@{}}
\toprule
\textbf{Layer} & \textbf{Dimension} & \textbf{Components} \\
\midrule
Input            & 1568 & -- \\
Hidden~1         & 1024 & Dense + BN + ReLU + Dropout(0.5) \\
Hidden~2         & 512  & Dense + BN + ReLU + Dropout(0.4) \\
Hidden~3         & 256  & Dense + BN + ReLU + Dropout(0.3) \\
Output (MNIST)   & 10   & Dense + Softmax \\
Output (Letters) & 26   & Dense + Softmax \\
\bottomrule
\end{tabular}
\end{table}

\paragraph{Parameter count.}

Including Dense biases and BN \(\gamma,\beta\), the trainable parameter counts are
\[
\begin{aligned}
\text{MNIST (10-way)}
&:~2{,}268{,}938
~\left(\approx 8.66~\mathrm{MB}\ \text{in FP32}\right),\\
\text{Letters (26-way)}
&:~2{,}273{,}050
~\left(\approx 8.67~\mathrm{MB}\ \text{in FP32}\right).
\end{aligned}
\]

These counts make the classifier footprint explicit. The simplicity of the proposed baseline lies primarily in its fixed two-channel representation and absence of learned spatial filtering; the dense classifier itself still contains approximately 2.27 million trainable parameters.

\subsection{Optimization and Training}

The model is implemented in TensorFlow/Keras. Optimization uses Adam with Keras defaults and sparse categorical cross-entropy. The batch size is 128 and training is allowed to continue for up to 50 epochs. Early stopping monitors validation accuracy with patience 4 and restores the best weights. A \texttt{ReduceLROnPlateau} callback monitors validation loss, halves the learning rate after three non-improving epochs, and uses a minimum learning rate of \(10^{-6}\). No explicit \(L_2\) weight decay or data augmentation is used.

In the executed notebooks, early stopping restored epoch 22 for MNIST and epoch 33 for EMNIST Letters.

\section{Experimental Setup}
\label{sec:setup}

\subsection{Datasets and Split Protocol}

MNIST~\cite{LeCun1998} is loaded from TensorFlow Datasets (TFDS) and contains 60{,}000 training images and 10{,}000 test images. The two original TFDS partitions are concatenated, producing 70{,}000 images, and are then repartitioned with a stratified 80/20 split using \texttt{train\_test\_split} with \texttt{random\_state=42}. This yields 56{,}000 images in the training portion and 14{,}000 in the test portion. During model fitting, Keras reserves 10\% of the training portion for validation, giving 50{,}400 fitting examples and 5{,}600 validation examples.

EMNIST Letters~\cite{Cohen2017} is loaded with the TFDS \texttt{emnist/letters} configuration used by the supplied notebook. In TFDS version 3.1.0, this configuration contains 88{,}800 training examples and 14{,}800 test examples, for 103{,}600 available examples in total~\cite{TensorFlowDatasets}. As with MNIST, the two TFDS partitions are concatenated and then repartitioned with the same stratified 80/20 procedure. The resulting training portion contains 82{,}880 examples and the test set contains 20{,}720. The 10\% validation split used during fitting leaves 74{,}592 training examples and 8{,}288 validation examples.

Table~\ref{tab:sizes} summarizes the exact sample counts used by the executed notebooks.

\begin{table}[htbp]
\centering
\caption{Dataset sizes under the executed preprocessing and split protocol. ``Train (pre-val)'' is the 80\% training partition before Keras reserves 10\% for validation.}
\label{tab:sizes}
\begin{tabular}{@{}lrrrrr@{}}
\toprule
\textbf{Dataset} & \textbf{Total} & \textbf{Train (pre-val)} & \textbf{Train (fit)} & \textbf{Val} & \textbf{Test} \\
\midrule
MNIST & 70{,}000 & 56{,}000 & 50{,}400 & 5{,}600 & 14{,}000 \\
TFDS EMNIST Letters & 103{,}600 & 82{,}880 & 74{,}592 & 8{,}288 & 20{,}720 \\
\bottomrule
\end{tabular}
\end{table}

\subsection{Label Handling}

MNIST uses labels 0--9 directly. In the EMNIST notebook, TFDS labels are shifted by subtracting one so that the 26 letter classes are represented by indices 0--25, corresponding to A--Z.

\subsection{Evaluation Metrics}

Evaluation uses the held-out test partition produced by the split procedure above. Top-1 accuracy is reported together with macro- and micro-averaged precision, recall, and F1. Macro averaging gives equal weight to each class, while micro averaging aggregates decisions over all test examples.

For probability-based evaluation, one-vs-rest ROC curves are computed from the softmax output for every class. The micro-average ROC is obtained by pooling all one-vs-rest decisions, while the macro-average ROC is obtained by interpolating classwise true-positive rates over a common false-positive-rate grid and averaging them. Confusion matrices report raw counts, with rows corresponding to true classes and columns to predicted classes.

\subsection{Software Environment}

Sobel derivatives are computed with OpenCV using
\texttt{cv2.Sobel($\cdot$,$\cdot$,1,0,ksize=3)}
and
\texttt{cv2.Sobel($\cdot$,$\cdot$,0,1,ksize=3)}.
The remaining pipeline uses TensorFlow/Keras, TensorFlow Datasets, NumPy, and scikit-learn.

\section{Results}
\label{sec:results}

\subsection{Aggregate Classification Metrics}

The executed MNIST model achieves \textbf{98.54\%} test accuracy. Macro precision, recall, and F1 are all 0.9853, while micro precision, recall, and F1 are 0.9854. The EMNIST Letters model achieves \textbf{92.50\%} test accuracy, with macro precision 0.9273, macro recall 0.9264, and macro F1 0.9265. Its micro precision, recall, and F1 are all 0.9250.

\begin{table}[htbp]
\centering
\caption{Aggregate classification metrics from the executed notebooks.}
\label{tab:aggregate_metrics}
\begin{tabular}{@{}lccccc@{}}
\toprule
\textbf{Dataset} & \textbf{Accuracy} & \textbf{Macro P} & \textbf{Macro R} & \textbf{Macro F1} & \textbf{Micro F1} \\
\midrule
MNIST & \textbf{98.54\%} & 0.9853 & 0.9853 & 0.9853 & 0.9854 \\
EMNIST Letters & \textbf{92.50\%} & 0.9273 & 0.9264 & 0.9265 & 0.9250 \\
\bottomrule
\end{tabular}
\end{table}

\subsection{Confusion-Matrix Analysis: MNIST}
\label{sec:mnist_cm}

\begin{figure}[htbp]
\centering
\includegraphics[width=0.90\linewidth]{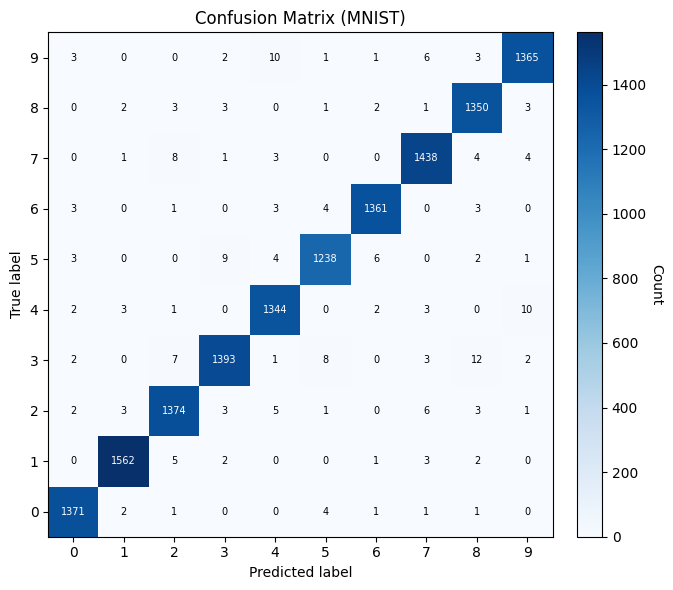}
\caption{Confusion matrix on the MNIST test set. Rows are true labels, columns are predicted labels, and entries are raw counts.}
\label{fig:cm_mnist}
\end{figure}

Figure~\ref{fig:cm_mnist} shows the MNIST confusion matrix. The matrix is strongly diagonal, consistent with the 98.54\% top-1 accuracy. Most off-diagonal entries are small relative to the class totals, indicating that the Sobel-gradient representation provides stable discrimination across all ten digits.

\noindent\textit{Notation.}
We write \(i\!\to\!j:c\) to indicate that \(c\) test examples with true class \(i\) were predicted as class \(j\).

The largest visible off-diagonal errors include \(3\!\to\!8:12\), \(4\!\to\!9:10\), \(9\!\to\!4:10\), \(5\!\to\!3:9\), and \(7\!\to\!2:8\). These errors are geometrically plausible. Handwritten 3 and 8 can become similar when the central opening or stroke separation is weak, while 4 and 9 can share an upper closed or partially closed structure. The reciprocal 4/9 confusion is especially informative because it suggests that first-order edge maps capture the overall contour well but can remain ambiguous when a small internal or terminal stroke determines the label.

\subsection{Confusion-Matrix Analysis: EMNIST Letters}
\label{sec:emnist_cm}

\begin{figure}[htbp]
\centering
\includegraphics[width=0.94\linewidth]{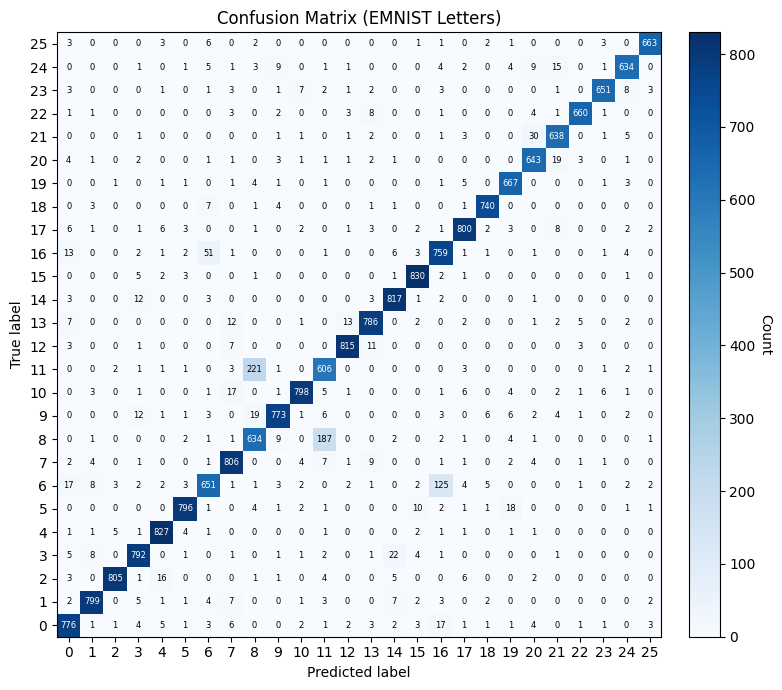}
\caption{Confusion matrix on the EMNIST Letters test set. Classes use the 0-based A--Z indexing employed by the notebook. Entries are raw counts.}
\label{fig:cm_emnist}
\end{figure}

Figure~\ref{fig:cm_emnist} shows the 26-class EMNIST Letters confusion matrix. Diagonal dominance remains strong, but the off-diagonal structure is more pronounced than for MNIST, which is consistent with both the larger number of classes and the greater visual similarity among handwritten letters.

\noindent\textit{A--Z index mapping.}\\
\noindent\small
A(0), B(1), C(2), D(3), E(4), F(5), G(6), H(7), I(8), J(9), K(10), L(11), M(12),
N(13), O(14), P(15), Q(16), R(17), S(18), T(19), U(20), V(21), W(22), X(23), Y(24), Z(25).
\normalsize

The strongest confusion is the I/L pair: \(I\!\to\!L:187\) and \(L\!\to\!I:221\). In many handwritten examples, both letters are dominated by a vertical stroke and differ only through short horizontal or terminal marks. Those small structures occupy relatively few pixels, so their contribution can be weak compared with the shared vertical-edge pattern.

A second notable pair is G/Q: \(G\!\to\!Q:125\) and \(Q\!\to\!G:51\). Both letters contain a rounded main body; the distinction often depends on a short cross-stroke or tail. The concentration of errors in these pairs suggests that the limitation is not a general failure of the representation, but a localized ambiguity when the class-defining feature is small relative to the dominant contour.

\subsection{ROC--AUC Analysis}
\label{sec:roc_analysis}

\begin{figure}[htbp]
\centering
\includegraphics[width=0.90\linewidth]{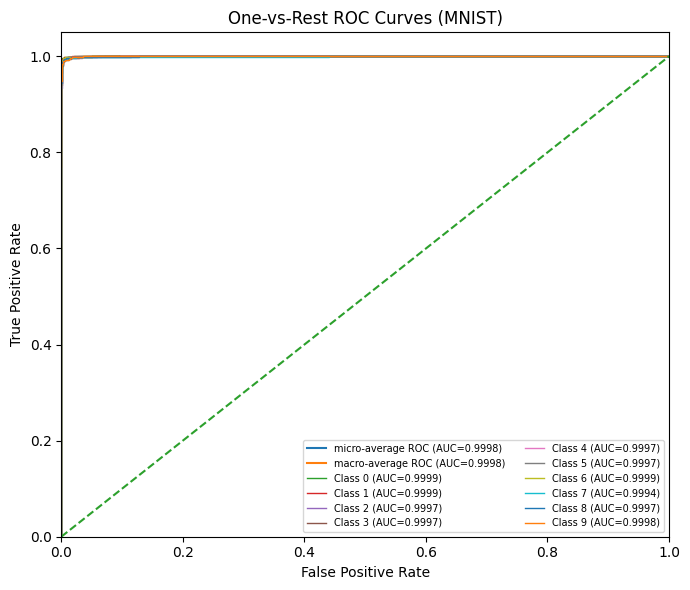}
\caption{One-vs-rest ROC curves for MNIST. Micro-average AUC \(=0.9998\) and macro-average AUC \(=0.9998\).}
\label{fig:roc_mnist}
\end{figure}

Figure~\ref{fig:roc_mnist} shows the one-vs-rest ROC curves for MNIST. The micro- and macro-average AUC values are both approximately 0.9998. The individual class AUCs are also tightly concentrated near one; the lowest displayed class AUC is approximately 0.9994. This indicates that the model's class scores provide extremely strong ranking separation between each digit and the remaining classes.

\begin{figure}[htbp]
\centering
\includegraphics[width=0.94\linewidth]{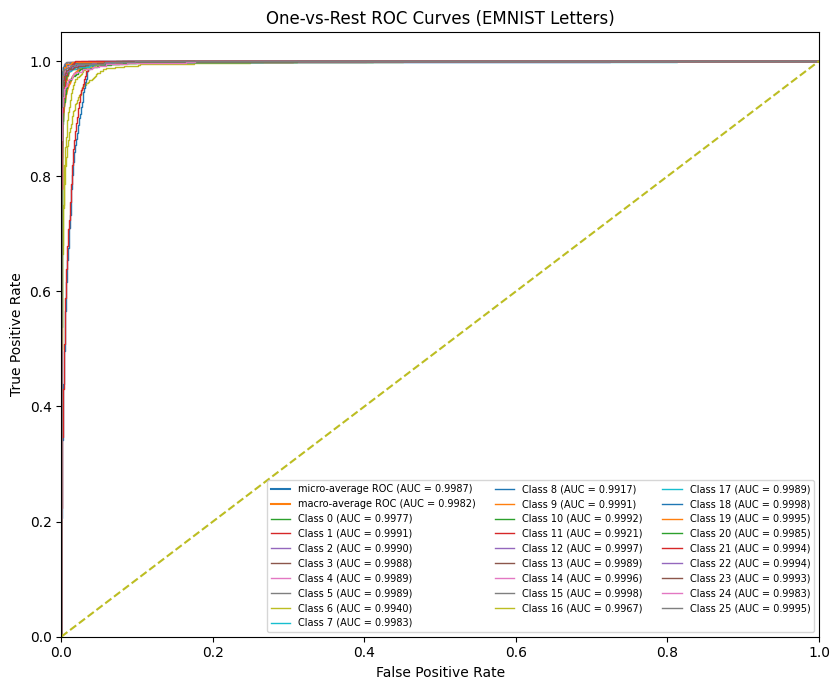}
\caption{One-vs-rest ROC curves for EMNIST Letters. Micro-average AUC \(=0.9987\) and macro-average AUC \(=0.9982\).}
\label{fig:roc_emnist}
\end{figure}

Figure~\ref{fig:roc_emnist} presents the corresponding analysis for EMNIST Letters. The micro-average AUC is 0.9987 and the macro-average AUC is 0.9982. Most classes exceed 0.998 AUC, while the lowest values occur for class 8 (I, 0.9917), class 11 (L, 0.9921), class 6 (G, 0.9940), and class 16 (Q, 0.9967). These are the same classes that participate in the most prominent confusion-matrix errors, providing consistent evidence that the main weaknesses arise from a small set of highly similar letter shapes.

The very high one-vs-rest AUC values should not be interpreted as equivalent to top-1 accuracy. AUC measures how well the score for a given class ranks positive examples above examples from all other classes. A sample can therefore contribute to a high one-vs-rest AUC even when a closely related competing class receives a slightly larger softmax score and wins the final argmax decision. This explains why EMNIST can exhibit micro/macro AUC values near 0.999 while its top-1 accuracy remains 92.50\%.

\FloatBarrier

\subsection{Error Patterns Across the Two Datasets}
\label{sec:error_patterns}

Taken together, the confusion matrices and ROC curves show two complementary properties of the Sobel-gradient representation. First, the overall ranking structure is strong: both datasets exhibit ROC curves concentrated near the upper-left corner. Second, the residual classification errors are highly structured rather than uniformly distributed. They occur predominantly between classes whose global edge layouts are similar and whose differences depend on a small hook, cross-stroke, terminal, opening, or internal segment.

This pattern is consistent with the inductive bias of the representation. Signed first-order gradients preserve the position, direction, and polarity of stroke boundaries, but they do not explicitly encode higher-order curvature, connectivity, topology, or stroke continuity. When two classes differ mainly through one of those properties, the flattened Sobel representation can leave the MLP with only a weak local cue.

\section{Discussion}
\label{sec:discussion}

\subsection{What the Results Establish}

The central finding is a representation-sufficiency result. Under the stated preprocessing and training protocol, a classifier that receives only two fixed signed Sobel derivative channels achieves 98.54\% accuracy on MNIST and 92.50\% on the TFDS EMNIST Letters configuration. The corresponding macro F1 and ROC--AUC results show that the performance is not explained solely by a small subset of classes: classification quality remains strong across the label space.

This conclusion is deliberately narrower than claiming that Sobel preprocessing is better than raw pixels or that the model outperforms CNNs. The present experiments do not include a matched raw-pixel MLP, a matched gradient-magnitude baseline, or a CNN trained under the identical split and optimization protocol. The results therefore establish what is achievable from the Sobel representation, not the incremental gain caused by Sobel preprocessing relative to alternative inputs.

\subsection{What the Confusion Matrices Reveal}

The confusion matrices provide a more informative picture than aggregate accuracy alone. On MNIST, errors are sparse and distributed among a few visually related digit pairs. On EMNIST Letters, the strongest errors concentrate around I/L and G/Q. These pairs share most of their coarse contour geometry and differ through relatively small structures. The result supports the interpretation that first-order edges capture the dominant glyph shape effectively but are less decisive when recognition depends on a short internal or terminal stroke.

This observation also suggests a concrete direction for extending the representation. Features that explicitly encode curvature, connectivity, or multiscale local structure may be most useful not across all classes uniformly, but specifically for resolving the pairs that remain ambiguous under first-order gradients.

\subsection{Interpreting the ROC Results}

The ROC curves show that class scores are highly separable even for EMNIST Letters. The difference between near-unity one-vs-rest AUC and lower top-1 accuracy is not contradictory: multiclass top-1 classification depends on the relative ordering among all competing classes for each sample, whereas one-vs-rest AUC asks whether a class score tends to rank its own examples above the pooled negatives. A model can therefore rank I examples well against most non-I letters while still occasionally assigning an even higher score to L. This behavior is visible in the combination of high AUC and concentrated I/L confusion.

\subsection{Why Signed First-Order Gradients Are Informative}

A handwritten glyph is largely defined by the arrangement of its stroke boundaries. The pair \((G_x,G_y)\) preserves both the location and directional polarity of local intensity changes. Flattening removes explicit local neighborhood processing inside the classifier, but it does not remove spatial identity: every feature dimension still corresponds to a fixed pixel location and derivative direction. The MLP can therefore learn global combinations of edge responses associated with loops, stems, crossbars, openings, and terminals.

The use of signed derivatives also differs from supplying only gradient magnitude. Positive and negative responses remain distinguishable, allowing the classifier to exploit the direction of light-to-dark and dark-to-light transitions. Whether this polarity is necessary for the reported performance remains an open empirical question; a magnitude-only and magnitude/orientation ablation would be required to answer it.

\subsection{Representation Simplicity versus Model Size}

The front end is computationally simple: two fixed \(3\times3\) derivative filters produce a 1568-dimensional vector. However, the MLP contains approximately 2.27 million trainable parameters. It is therefore more precise to describe the method as having a simple, fixed, and interpretable \emph{representation} rather than to assume that the entire system is smaller than arbitrary CNN alternatives. Claims about memory, latency, or energy advantages require matched implementations and direct measurements.

\subsection{Rotation, Alignment, and Normalization}

The representation is not rotation-invariant. Rotating a glyph changes how edge energy is distributed between \(G_x\) and \(G_y\), while flattening ties each response to an absolute image location. Off-centering can similarly shift informative features to coordinates that the dense classifier has not learned to associate with the class. Data augmentation, alignment procedures, or rotation-aware derivative encodings could reduce these sensitivities.

Independent per-image min--max normalization introduces a second trade-off. It standardizes numerical scale and emphasizes the relative pattern of edges within each glyph, but it also discards absolute differences in gradient amplitude between images. A future ablation comparing per-image normalization, dataset-level normalization, and unnormalized signed derivatives would clarify whether this choice helps or limits generalization.

\subsection{Limitations and Threats to Validity}

Several limitations constrain the interpretation of the reported results.

\begin{enumerate}[leftmargin=1.5em]

    \item \textbf{Study-specific re-splitting.}
    The original TFDS train and test partitions are concatenated before the stratified 80/20 split. The resulting test sets are therefore study-specific and should not be treated as identical to canonical benchmark test partitions.

    \item \textbf{TFDS-specific EMNIST configuration.}
    The executed notebook uses TFDS \texttt{emnist/letters} version 3.1.0, whose available split sizes differ from the full EMNIST Letters counts often quoted from the original NIST release. Comparisons should therefore use the same TFDS configuration and preprocessing protocol.

    \item \textbf{Single-run reporting.}
    The reported metrics come from one training run for each dataset. Variation across random initializations is not measured, and no confidence intervals are reported.

    \item \textbf{No matched input ablations.}
    A raw-pixel MLP, magnitude-only gradient MLP, and magnitude/orientation MLP are not trained under the same protocol. The study demonstrates the sufficiency of signed Sobel gradients but does not quantify their causal advantage over alternative input representations.

    \item \textbf{No matched CNN baseline.}
    A CNN is not trained using the same re-split, optimization budget, and evaluation procedure. The present results therefore should not be interpreted as a direct CNN comparison.

    \item \textbf{No robustness evaluation.}
    Noise, blur, rotation, translation, and stroke-thickness perturbations are not systematically evaluated.

    \item \textbf{No deployment benchmark.}
    Although Sobel filtering itself is inexpensive, the notebooks do not provide a controlled hardware benchmark for end-to-end latency, memory, or energy. The recorded wall-clock times are environment-dependent and are therefore not used as deployment evidence here.

\end{enumerate}

\section{Conclusion and Future Work}
\label{sec:conclusion}

This study evaluates a deliberately constrained handwritten-character recognition pipeline in which the visual representation is limited to fixed first-order Sobel derivatives and the learned component is an all-dense MLP. The executed models attain 98.54\% test accuracy on MNIST and 92.50\% on the TFDS EMNIST Letters configuration, with macro F1 scores of 0.9853 and 0.9265, respectively. One-vs-rest ROC analysis yields micro/macro AUC values of 0.9998/0.9998 on MNIST and 0.9987/0.9982 on EMNIST Letters.

The confusion matrices show that the remaining errors are concentrated among a small number of geometrically similar classes rather than being broadly distributed. In particular, the I/L and G/Q confusions in EMNIST demonstrate the main limitation of the first-order representation: when two classes share nearly the same dominant contour and differ only through a short local stroke, signed edge maps may provide insufficient separation for a flattened MLP.

The result is therefore best interpreted as a strong baseline about representation sufficiency rather than as evidence that Sobel gradients are superior to raw pixels or learned CNN features. The next experiments should compare Sobel inputs directly with raw pixels, gradient magnitude/orientation, and a matched CNN under the same split and optimization budget; repeat training across multiple random seeds; evaluate robustness to rotation, translation, noise, blur, and stroke-thickness variation; and test whether adding higher-order geometric information specifically reduces the class-pair ambiguities identified by the confusion matrices. These experiments would clarify not only how well fixed first-order gradients work, but also exactly where more expressive spatial representations become necessary.

\section*{Code Availability}

All code and training scripts are available at\\
\url{https://github.com/AzamNouri/Sobel-Gradient-MLP}.\\
See related articles and supporting materials in \cite{nouri2025ijca}
\cite{nouri2025critical}
\cite{nouri2026siggain}
\cite{nouri2026stepcache}
\cite{nouri2026mdl}.


\bibliographystyle{plain}

\end{document}